%% file: paper.tex
\documentclass{article}
\usepackage{spconf,amsmath,graphicx,hyperref}

\usepackage{multirow}
\usepackage{graphicx}
\usepackage{subfigure}

\title{EntroCut: Entropy-Guided Adaptive Truncation for Efficient Chain-of-Thought Reasoning in Small-scale Large Reasoning Models}
\name{Hongxi Yan$^{1,2}$, Qingjie Liu$^{1,2,3,\dag}$, Yunhong Wang$^{1}$\thanks{$\dag$ Corresponding Author}}
\address{
$^1$ Beihang University, School of Computer Science \& Engineering, China \\
$^2$ Zhongguancun Laboratory, China \\
$^3$ Hangzhou Innovation Institute, China \\
}
\begin{document}
% \ninept
%
\maketitle
\begin{abstract}
% Large Reasoning Models (LRMs) excel at complex reasoning tasks through extended chain-of-thought (CoT) generation, but their reliance on lengthy intermediate steps incurs substantial computational cost. We find that the entropy of the model’s response distribution in early reasoning steps reliably distinguishes correct from incorrect reasoning. Motivated by this observation, we propose EntroCut, a training-free method that dynamically truncates reasoning by identifying high-confidence states where reasoning can be safely terminated. Experiments on 4 benchmarks—AIME24, AIME25, Math500, and AMC23—show that EntroCut reduces token usage by up to 40\% while with minimal sacrifice in accuracy, achieving superior efficiency-performance trade-offs compared with existing training-free methods. These results demonstrate that entropy-guided dynamic truncation provides a practical and effective approach to mitigate the inefficiency of LRMs.
Large Reasoning Models (LRMs) excel at complex reasoning tasks through extended chain-of-thought generation, but their reliance on lengthy intermediate steps incurs substantial computational cost. We find that the entropy of the model's output distribution in early reasoning steps reliably distinguishes correct from incorrect reasoning. Motivated by this observation, we propose EntroCut, a training-free method that dynamically truncates reasoning by identifying high-confidence states where reasoning can be safely terminated. To comprehensively evaluate the trade-off between efficiency and accuracy, we introduce the Efficiency-Performance Ratio (EPR), a unified metric that quantifies relative token savings per unit accuracy loss. Experiments on four benchmarks show that EntroCut reduces token usage by up to 40\% with minimal accuracy sacrifice, achieving superior efficiency-performance trade-offs compared with existing training-free methods. These results demonstrate that entropy-guided dynamic truncation provides a practical approach to mitigate the inefficiency of LRMs.
\end{abstract}
\section{Introduction}
\label{sec:intro}

Large Reasoning Models (LRMs) have revolutionized complex reasoning tasks such as mathematics and programming through their ability to generate extended Chain-of-Thought (CoT) sequences \cite{deepseekai2025deepseekr1incentivizingreasoningcapability, wei2022chain}. However, this reliance on extended reasoning processes, which are characterized by lengthy intermediate steps, creates a fundamental efficiency paradox. While such processes are essential for tackling intricate problems, their inherent computational intensity leads to significant inefficiencies \cite{sui2025stopoverthinkingsurveyefficient}. Achieving a balanced CoT reasoning length while maintaining high accuracy has emerged as a crucial and promising strategy, warranting deeper exploration \cite{feng2025efficientreasoningmodelssurvey}. 

A variety of training-free methods have emerged to balance reasoning length and model performance. Compared with SFT- \cite{ma-etal-2025-cot,liu2024can,han-etal-2025-token} or RL-based \cite{yeo2025demystifying,aggarwal2025l,luo2025opruner} strategies, training-free approaches do not necessitate large-scale data or extensive computational resources, and they are easily deployable and broadly compatible, avoiding the necessity of retraining for different models. Existing training-free techniques mainly focus on suppression of reflection ability \cite{nowait, tip} or applying fixed truncation rules \cite{muennighoff2025s1}. However, aggressive suppression of reflection ability may disrupt the model’s reasoning patterns, a problem especially severe for small-scale models with limited generalization capacity \cite{nowait}. Moreover, rigid truncation may prematurely cut off valid reasoning or fail to prevent unnecessary continuation, leading to inefficient allocation of reasoning resources.

%However, aggressive suppression of reflection ability may disrupt the model’s reasoning patterns and degrade accuracy—particularly for lower-parameter versions of the model—while rigid truncation can prematurely cut off valid reasoning or fail to prevent unnecessary continuation, effects that such small models are generally unable to compensate for \cite{nowait}.

\begin{figure}[!t]
    \centering
    \subfigure[Response Entropy Distribution]{
        \includegraphics[width=0.22\textwidth]{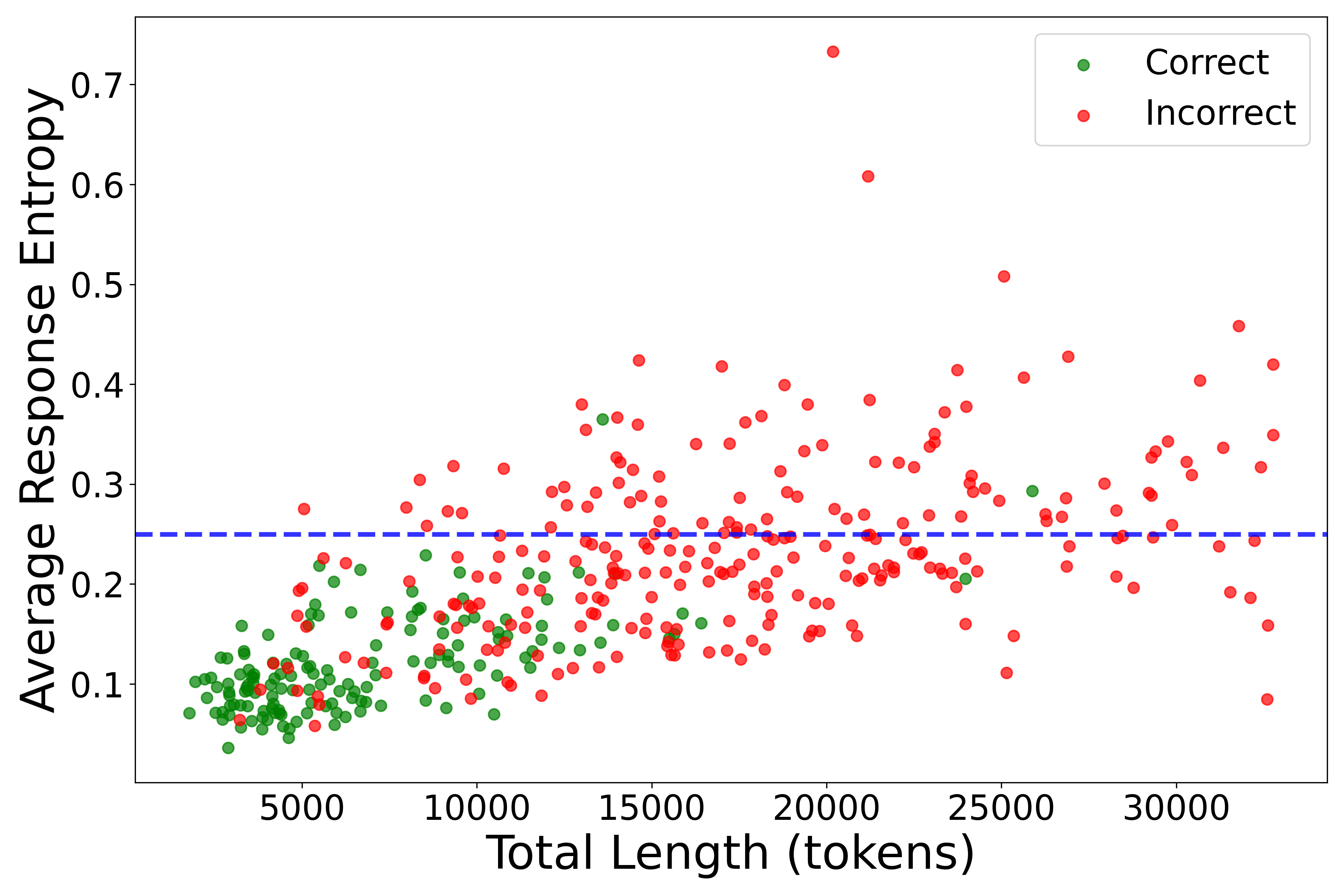}
        \label{fig:average_entropy}
    }
    \hfill
    \subfigure[Prefix Mean Entropy Curve]{
        \includegraphics[width=0.22\textwidth]{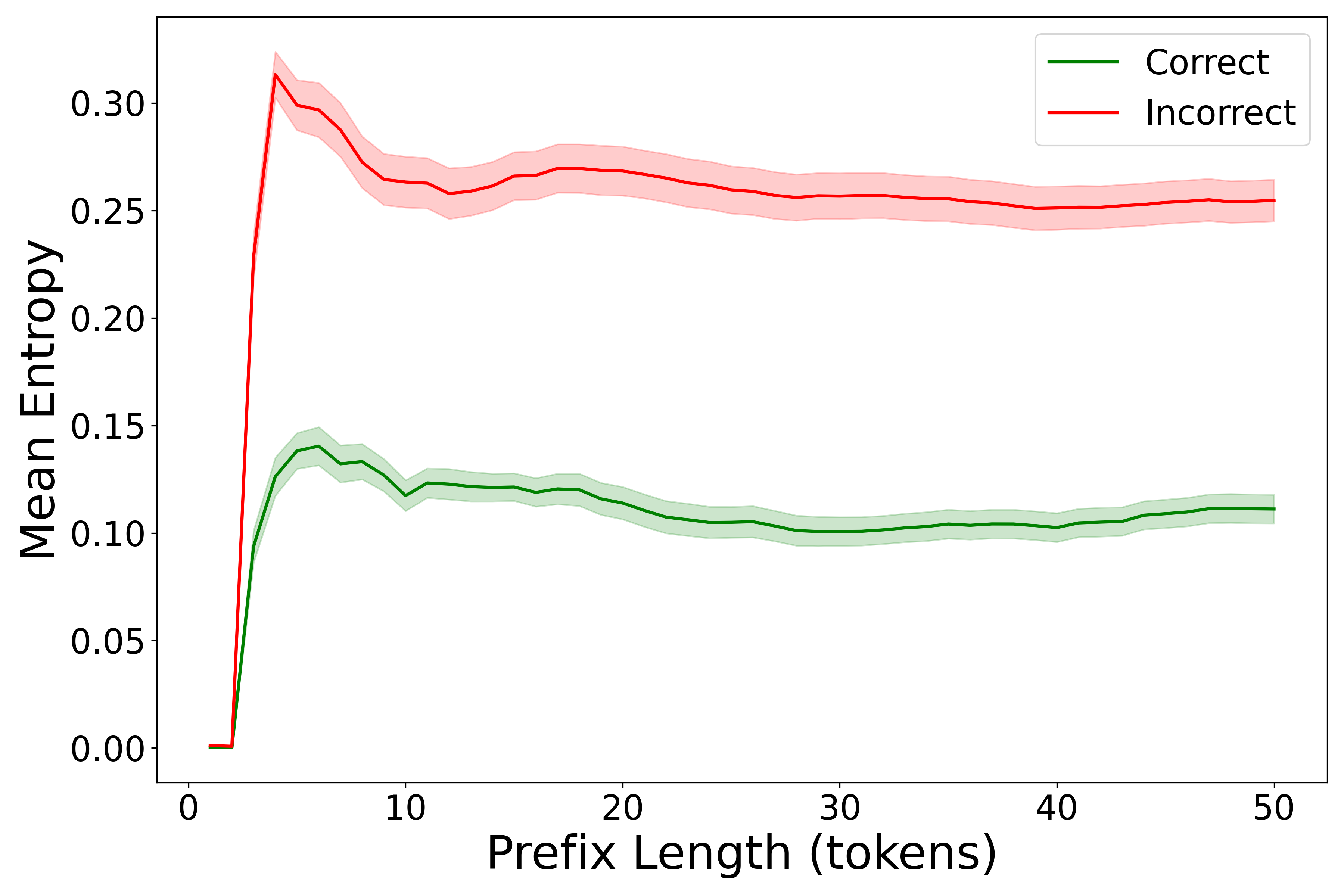}
        \label{fig:prefix_entropy}
    }
    \caption{Entropy Analysis of Deepseek-Distill-1.5B (AIME24)}
    \label{fig:entropy picture}
\vspace{-0.5em}
\end{figure}

Our investigation reveals a promising solution: the entropy of the model's response distribution during its content generation phase exhibits a distinct separation between correct and incorrect samples, as visualized in Figure~\ref{fig:average_entropy}. Crucially, this distinguishing entropy signal emerges early in the generation process (as shown in Figure~\ref{fig:prefix_entropy}), requiring only a few tokens to be computed. Motivated by this finding, we propose EntroCut, a training-free approach that probes the prefix entropy of responses to dynamically decide whether to terminate or continue reasoning. This entropy-guided mechanism prevents excessive reasoning beyond necessity, striking a balance between computational efficiency and response accuracy. In summary, our contributions are as follows:

\textbf{Entropy as a Confidence Signal \& EntroCut Algorithm}: We demonstrate for the first time that correct and incorrect responses diverge clearly in early-step entropy, providing a reliable signal for truncating or continuing reasoning. Based on this insight, we introduce EntroCut, a training-free strategy that adaptively terminates reasoning once sufficient confidence is reached, reducing unnecessary computation while maintaining accuracy.

\textbf{A Principled Measure of Efficiency–Accuracy Trade-off}: To better capture the trade-off between efficiency and performance, we propose the Efficiency-Performance Ratio (EPR). Anchored by the full reasoning and minimal reasoning baselines, EPR quantifies relative token savings per unit accuracy loss, providing a unified and interpretable measure of reasoning efficiency.

\textbf{Improved Efficiency with Minimal Accuracy Sacrifice}: Extensive experiments show that EntroCut consistently shortens reasoning chains and reduces latency, while maintaining accuracy or incurring only minimal loss. This strikes an optimal balance between efficiency and performance for LRMs.

\section{Method}
\label{sec:method}

\begin{figure}[htb]
\includegraphics[width=0.48\textwidth]{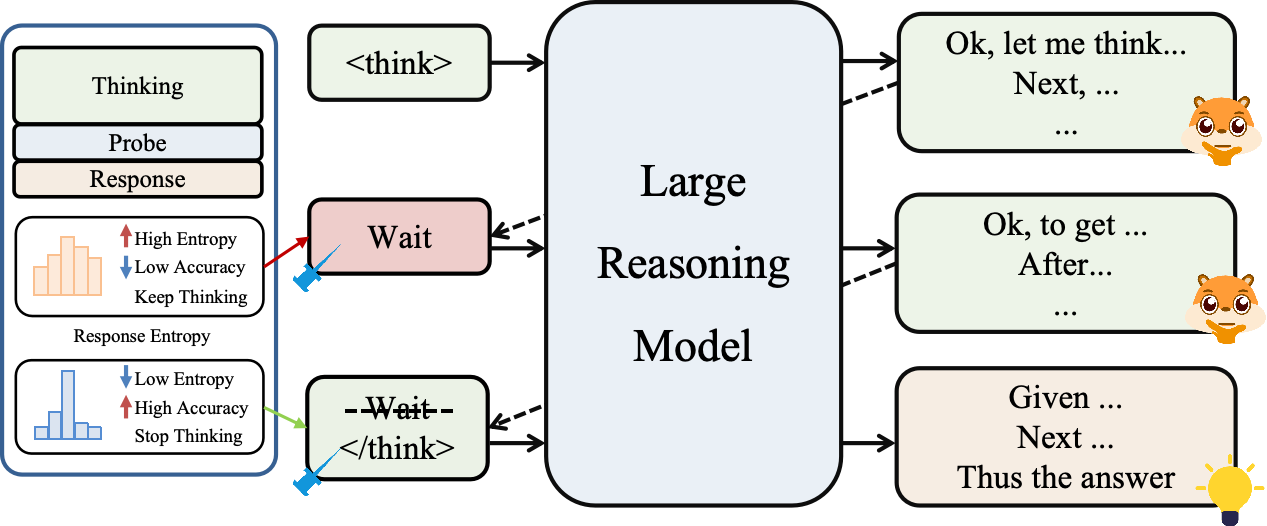}
\caption{The overview of EntroCut.}
\label{fig:overview}
\end{figure}

We propose \textbf{EntroCut} (Fig.~\ref{fig:overview}), an entropy-guided framework that adaptively prunes reasoning chains of LRMs. The key insight is that token-level entropy reflects model confusion, and correct versus incorrect responses exhibit distinct entropy patterns in the early steps. By probing this signal and terminating once confidence is detected, EntroCut reduces reasoning length while maintaining performance.

\subsection{Reasoning paradigm}

LRMs employ a two-stage reasoning process to solve complex problems, as observed in models exhibiting CoT capabilities \cite{li202512surveyreasoning}. This process is typically delineated into a reflective reasoning phase and a final response generation phase which can be formally defined as follows:

\textbf{Chain-of-Thought Reasoning Phase:} The model iteratively generates reasoning steps to decompose and reflect on the problem until reasoning stopping condition is met.
\vspace{-0.5em}
$$\mathbf{t}_k = \pi_{\theta}(\mathbf{t}_k \mid \mathbf{q}, \mathbf{t}_{<k}) \quad \text{for } k=1, 2, \ldots, K$$

Here, $\pi_{\theta}$ represents the large reasoning model with parameters $\theta$, $\mathbf{q}$ represents the user query, and $\mathbf{t}_k$ is the $k$-th reasoning token. The process terminates at thinking stop token (\texttt{</think>}), denoted as $\mathbf{t}_{\text{stop}} = \mathbf{t}_K$.

\textbf{Final Response Generation Phase:} Following the internal reasoning, the model synthesizes a coherent and concise answer based on the thought process.
\vspace{-0.5em}
$$\mathbf{r}_k = \pi_{\theta}(\mathbf{r}_k \mid \mathbf{q}, \mathbf{t}_{1:K}, \mathbf{r}_{<k}) \quad \text{for } k=1, 2, \ldots, L$$

Here, $\mathbf{r}_k$ represents the $k$-th token of the final response  by the model $\pi_{\theta}$, and $\mathbf{t}_{1:K}$ denotes the full sequence of reasoning tokens.

\subsection{Entropy based adaptive think budget}

% While Chain-of-Thought Reasoning Phase allows for in-depth problem analysis, it often leads to verbose outputs and high inference latency due to the generation of redundant self-reflections and turns of phrase (e.g., "wait," "but"). Conventional hard budgeting method \cite{muennighoff2025s1} attempt to address this by prematurely truncating the reasoning phase, but this risks losing critical intermediate steps needed for a high-quality final answer.

LRMs' reasoning phase often becomes verbose due to excessive self-reflection with transition tokens (e.g., ``wait", ``but"); while hard-budgeting approaches mitigate this by truncating the thinking phase \cite{muennighoff2025s1}, they overlook the fundamental fact that different problems demand different cognitive budgets. Resulting in suboptimal resource allocation and degraded performance across tasks of varying complexity.

% To overcome this limitation, we introduce the Adaptive Think Budget mechanism. Our approach enforces an early termination of the explicit chain-of-thought phase, proactively reducing the generation of redundant or overly reflective reasoning tokens.

% Unlike hard budgeting, however, Adaptive Think Budget does not force the model to immediately output the final answer. Instead, it allows the model to transition to the content generation phase earlier than usual. This is a critical distinction: the model can then continue its problem-solving process implicitly during the response generation, using the more direct and structured nature of the final output to complete any reasoning steps that were prematurely truncated. This ensures that the model's final response is not only concise and efficient but also logically complete, as the model can implicitly synthesize truncated intermediate steps directly into the final content.

To overcome this limitation, we introduce the Entropy based Adaptive Think Budget mechanism. Our approach leverages entropy signals to dynamically determine whether the model has performed sufficient reasoning, and accordingly applies early termination to the reasoning phase. Moreover, the model is not required to output the final answer immediately after truncation. Instead, it is allowed to continue problem solving in the response generation phase. Compared with the reasoning stage, the response phase involves fewer digressions and transitions, enabling the model to produce results more efficiently while compensating for potential discontinuities caused by premature termination of reasoning.

% This method effectively balances efficiency and performance by eliminating the overhead of explicit, reflective reasoning while leveraging the coherence of the content generation phase to maintain a complete and accurate solution.

\subsection{Entropy probe for adaptive thinking}

A key challenge in implementing an adaptive reasoning budget is determining the optimal moment to transition from reasoning phase to generation phase. Our core hypothesis is that a model's confidence in its internal reasoning state can be measured by its uncertainty, which is quantifiable through the entropy of its predicted tokens in the generation phase.

To measure this confidence, we append a special probe string (e.g. \texttt{</think>/n/nSo the final answer is}) to the current sequence of reasoning tokens. The token-level entropy is then calculated as:

$$
\begin{aligned}
H_i = - \sum_{j} & p(r_j \mid \mathbf{t}_{\text{current}}, \text{probe}, r_{<i}) \\
& \cdot \log p(r_j \mid \mathbf{t}_{\text{current}}, \text{probe}, r_{<i})
\end{aligned}
$$

\input{tables/main_result}

Here, $\mathbf{t}_{\text{current}}$ represents the current sequence of reasoning tokens, ``probe" is the appended probe string, and $r_{<i}$ are the tokens generated thus far in the following sequence. We then compute the average entropy of the $k$ tokens generated immediately following this probe:

$$\bar{H}_{\text{probe}} = \frac{1}{k} \sum_{i=1}^{k} H_i$$

The reasoning phase is terminated if the average entropy of these probe tokens falls below a predefined threshold $\tau$ (i.e., $\bar{H}_{\text{probe}} \le \tau$). This low entropy indicates high model confidence, signaling that the model has likely conducted sufficient reasoning and is ready to proceed with the final answer.

To reduce unnecessary computational overhead, we do not apply entropy probes at every token or at the end of each sentence. We recognize that LRMs typically produce a relatively complete segment of thought before entering a moment of reflection. Just before this reflection occurs, the model has typically reached a stable internal state, making this the ideal point to conduct an entropy probe. Based on this insight, we design a mechanism that detects specific reflective tokens during inference, such as ``Wait". Once such a token is identified, the probe is triggered to calculate the average entropy of the subsequent tokens. This targeted strategy ensures that entropy checks are both timely and efficient, while aligning with the natural reasoning dynamics of the model.

% This entropy-based mechanism allows Adaptive Think Budget to make informed decisions on when to early exit the reasoning phase, thereby significantly improving efficiency while maintaining the performance.

\section{Experiments}
\label{sec:exps}

\subsection{Experiment settings}

\textbf{Datasets:} We conducted experiments on four datasets: AIME24, AIME25 \cite{aime}, MATH500 \cite{hendrycks2021measuring}, and AMC23 \cite{amc}. The AIME and AIME2025 datasets cover a wide range of mathematical problems in algebra, combinatorics, number theory, and probability. The MATH500 dataset includes problems ranging from middle school to university level, while the AMC2023 dataset is specifically sourced from the AMC12 2023 competition, which focuses on more fundamental mathematical reasoning problems.

\textbf{Metrics:} We evaluate the performance of our methods using Accuracy (Acc), Average Token Count (Tok), and the Efficiency-Performance Ratio (EPR). Acc measures the correctness of the final answer, while Tok assesses the computational cost by representing the average number of generated tokens per problem. To provide a unified and comprehensive assessment of a method's efficiency by quantifying its computational return for a given performance loss, we propose the EPR. It is defined as the ratio of relative token savings to relative accuracy loss:

$$EPR = \frac{\text{Token Saving Ratio}}{\text{Accuracy Loss Ratio}}.$$

% These ratios are calculated using two anchor points as a baseline: the Vanilla method, which represents the highest possible accuracy at maximum token cost, and the NOWAIT method, which achieves minimum token cost by suppressing all thought tokens. The Token Saving Ratio is the token reduction of the target method relative to the maximum possible reduction (from Vanilla to NOWAIT), defined as $\frac{\text{Vanilla\_Tok} - \text{Target\_Tok}}{\text{Vanilla\_Tok} - \text{NOWAIT\_Tok}}$. Similarly, the Accuracy Loss Ratio is the corresponding accuracy drop relative to the maximum possible loss, calculated as $\frac{\text{Vanilla\_Acc} - \text{Target\_Acc}}{\text{Vanilla\_Acc} - \text{NOWAIT\_Acc}}$. Here, $Target$ refers to the method being evaluated. A higher EPR value indicates a more effective balance between accuracy and token consumption.

These ratios are calculated using two anchor points as a baseline: the Vanilla method, which corresponds to the original reasoning process and represents the highest possible accuracy at maximum token cost, and the NOWAIT method, which achieves minimum token cost by suppressing all transition tokens. The Token Saving Ratio, measures the token reduction of the target method relative to the maximum possible reduction (from Vanilla to NOWAIT) which defined as

\vspace{-1em}
$$
\text{Token Saving Ratio} = \frac{\text{Vanilla\_Tok} - \text{Target\_Tok}}{\text{Vanilla\_Tok} - \text{NOWAIT\_Tok}},
$$

while the Accuracy Loss Ratio, measures the corresponding accuracy drop relative to the maximum possible loss:

$$
\text{Accuracy Loss Ratio} = \frac{\text{Vanilla\_Acc} - \text{Target\_Acc}}{\text{Vanilla\_Acc} - \text{NOWAIT\_Acc}}.
$$

Here, Target refers to the method being evaluated. A higher EPR value indicates a more effective balance between accuracy and token consumption.

Moreover, to ensure the reliability of our results, we performed multiple repeated tests for each dataset, specifically 16 repetitions for AIME, AIME2025, and AMC2023, and 4 repetitions for the MATH dataset. The final results are derived by averaging the outcomes of these tests.

\textbf{Models:} All experiments were performed on the DeepSeek-R1-Distill-Qwen-1.5B (DS-1.5B) and DeepSeek-R1-Distill-Qwen-7B (DS-7B).  For our tests, we set the temperature to 0.6 and top-p to 1.0. The entropy threshold $\tau$ was set to 0.15 for DS-1.5B (0.2 for AMC23) and 0.225 for DS-7B (0.15 for AMC23).
% \textbf{Models:} All experiments were performed on the DeepSeek-R1-Distill-Qwen-1.5B (DS-1.5B) and DeepSeek-R1-Distill-Qwen-7B (DS-7B).  For our tests, we set the temperature to 0.6 and top-p to 1.0. The entropy threshold $\tau$ was set to 0.15 for DS-1.5B (0.2 for AMC23) and 0.225 for DS-7B.

\subsection{Main results}

\begin{figure}[!t]
    \centering
    \subfigure[AIME24 Pareto Optimality]{
        \includegraphics[width=0.22\textwidth]{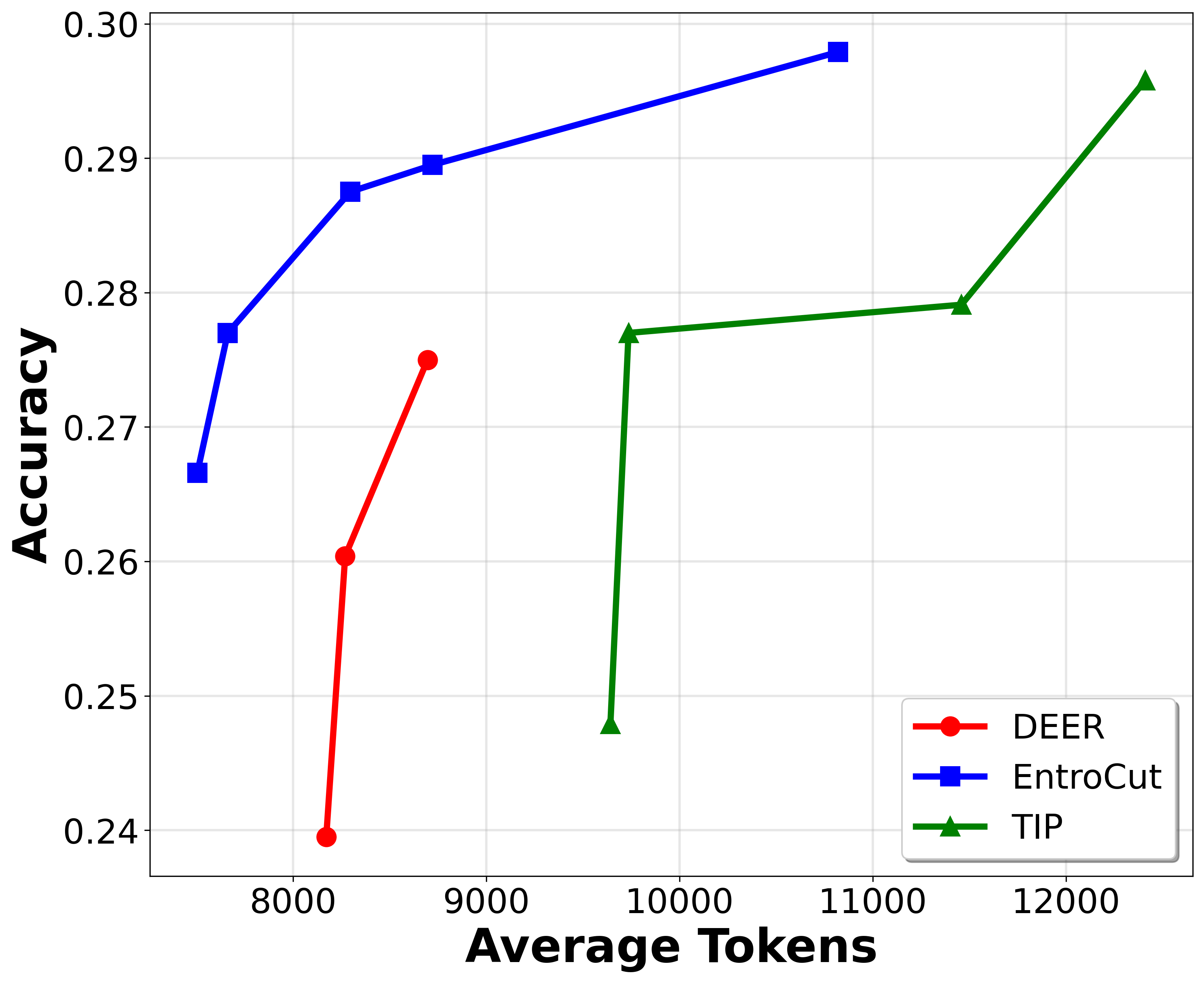}
    }
    \hfill
    \subfigure[AIME25 Pareto Optimality]{
        \includegraphics[width=0.22\textwidth]{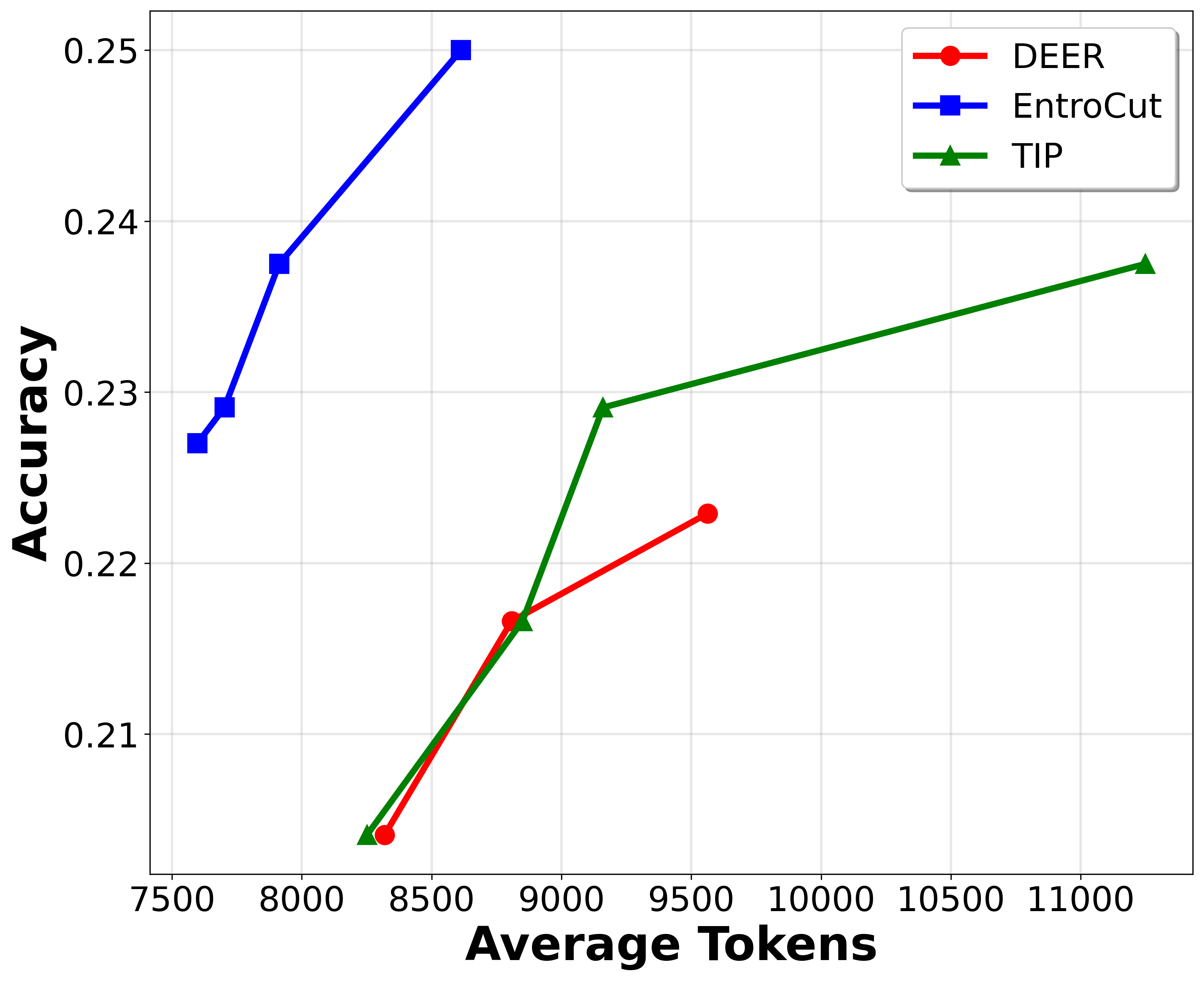}
    }
    % \hfill
    % \subfigure[Math500 Pareto Optimality]{
    %     \includegraphics[width=0.22\textwidth]{figures/plato_curves/monotonic_curves_math.png}
    % }
    % \hfill
    % \subfigure[AMC23 Pareto Optimality]{
    %     \includegraphics[width=0.22\textwidth]{figures/plato_curves/monotonic_curves_amc.png}
    % }
    \caption{DeepSeek-R1-Distill-Qwen-1.5B Pareto Optimality Curves on AIME24 and AIME25.}
    \label{fig:plato_curves}
\end{figure}

The main experimental results are presented in Table~\ref{tab:method_comparison_with_epr_compact}. The values in parentheses indicate the number of additional tokens introduced, which are negligible compared to the overall token savings. As shown in the table, our method achieves the highest accuracy while substantially reducing the number of tokens consumed. More importantly, it yields consistent improvements in EPR across all datasets, with average gains of 9.5 and 0.9 on DS-1.5B and DS-7B, respectively.

By comparison, the NOWAIT \cite{nowait} method entirely suppresses reflection tokens, forcing the model into a non-reasoning mode. Although this dramatically lowers token usage, it incurs a severe accuracy drop.  TIP \cite{tip}, which only slightly suppresses reflection tokens at the beginning of each reasoning stage, still disrupts the model’s natural reasoning process. This disruption is particularly detrimental on more challenging datasets such as AIME24 and AIME25, where deep reasoning is essential. DEER \cite{deer} employs the average answer probability during the reasoning phase as its truncation criterion. However, this not only requires a rigid output format that restricts generalizability but also relies on a metric less effective than entropy for capturing model uncertainty.

These comparisons highlight the advantages of our proposed EntroCut. By leveraging entropy as a robust uncertainty signal, our method avoids the degeneration of reasoning seen in NOWAIT and TIP, and achieves greater adaptability and effectiveness than DEER. As a result, EntroCut consistently delivers superior performance in accuracy and optimal balance between inference efficiency and model performance.

\subsection{Pareto optimality}

% 为了更好的证明EntroCut方法的有效性，我们对TIP，DEER和EntroCut方法选取的多组参数进行评测，并绘制出在四个数据集上tokens和Accuracy上的帕累托最优曲线，如图3所示。可以看到EntroCut在其包含的token范围内是最佳的。值得注意的是，在更困难的aime24和aime25上，EntroCut为全局的帕累托最优解，这有效的证明了EntroCut能够更好的应对困难问题，实现长序列思考推理的高效优化。

% To better demonstrate the effectiveness of our EntroCut method, we evaluated multiple parameter sets for the TIP, DEER, and EntroCut methods. We then plotted the Pareto optimal curves for tokens and accuracy of AIME24 and AIME25, as shown in Figure~\ref{fig:plato_curves}. The results indicate that EntroCut is the superior method within its token range. Notably, on the more challenging AIME24 and AIME25 datasets, EntroCut represents the global Pareto optimal solution. This effectively demonstrates that EntroCut is better suited to handling complex problems, achieving efficient optimization of long-sequence reasoning.

To demonstrate the effectiveness of our EntroCut method, we evaluated multiple parameter sets for TIP, DEER, and EntroCut. We plotted the Pareto optimal curves for tokens and accuracy of AIME24 and AIME25, as shown in Figure~\ref{fig:plato_curves}. The results indicate that EntroCut is the superior method within its token range. Notably, on the more challenging AIME24 and AIME25 datasets, EntroCut represents the global Pareto optimal solution. This demonstrates that EntroCut is better suited to handling complex problems, achieving efficient optimization of long-sequence reasoning.

\subsection{Ablation study}
\input{tables/ablation_study}

% 我们在AIME25上进行了消融实验。其中w/ hard budget表示当结束thinking后立刻让模型输出结果，w/o entropy probe表示使用固定长度的token结束思考，而不是动态根据entropy结束思考。如果直接输出结果，可以减少token但同样会降低accuracy，尤其是在能力更强的DS-7B上。这表明response阶段的进一步推理对于提早截断思考是很重要的。而使用固定的token长度截断思考则会降低准确率，这表明Entropy Probe是更好的识别合适应该截断而何时应该继续进行思考。

Table \ref{tab:entrocut_albation} presents ablation studies on AIME25. The ``w/ hard budget" variant, which forces the model to output a result immediately after the thinking phase, led to a sharp decrease in accuracy from 38.1\% to 17.2\% on the more capable DS-7B model, despite a reduction in token consumption. This indicates that further reasoning in the response phase is crucial for maintaining model performance when the thinking process is prematurely truncated. Another variant, ``w/o entropy probe", used a fixed token length to terminate thinking, resulting in a 1.1 percentage point drop in accuracy on the DS-7B model compared to the full Entrocut method. Furthermore, the EPR metric also decreased from 2.04 to 1.62 for DS-7B and significantly from 4.05 to 1.69 for DS-1.5B. This demonstrates that the Entropy Probe can intelligently determine when to stop thinking and provide an answer, achieving efficient thinking truncation with only minor accuracy loss, which makes it superior to a fixed-length strategy.

\section{Conclusion}
\label{sec:Conclusion}

% In this work, we proposed EntroCut, a training-free method that leverages entropy-guided adaptive truncation to balance reasoning efficiency and accuracy. By integrating entropy probes with a adaptive think budget, EntroCut eliminates redundant reasoning while preserving problem-solving capability in the response phase. Extensive experiments across multiple benchmarks demonstrate up to 40\% token reduction without sacrificing accuracy, achieving Pareto-optimal performance on AIME24 and AIME25. These results highlight entropy-guided truncation as a practical solution to the efficiency challenge in large reasoning models and a promising direction for future research on efficient inference.

% We propose EntroCut, a training-free method that leverages entropy-guided adaptive truncation to improve reasoning efficiency while maintaining accuracy. To evaluate efficiency in a unified way, we introduce EPR, a new metric quantifying the trade-off between token savings and accuracy loss. Experiments on multiple benchmarks show that EntroCut achieves substantial token reduction, minimal accuracy degradation and consistently high EPR, demonstrating its effectiveness and practical potential for efficient inference in large reasoning models.

We propose EntroCut, a training-free method that leverages entropy-guided adaptive truncation to improve reasoning efficiency while maintaining accuracy. To evaluate efficiency in a unified way, we introduce EPR, a new metric quantifying the trade-off between token savings and accuracy loss. Experiments on multiple benchmarks show that EntroCut achieves minimal accuracy degradation and consistently high EPR, demonstrating its effectiveness and practical potential for efficient inference in large reasoning models.

\vfill\pagebreak
\section{Acknowledgments}
\label{sec:Conclusion}
% This research was supported by ``Pioneer'' and ``Leading Goose'' R\&D Program of Zhejiang (No.~2024C01020); This research was supported by the Key R\&D Program of Zhejiang Province (No.~2024C01020); This research was supported by the Emergency Management Research and Development Project of Zhejiang Province (No.~2024YJ018).
This research was supported by ``Pioneer'' and ``Leading Goose'' R\&D Program of Zhejiang (No.~2024C01020);  the Key R\&D Program of Zhejiang Province (No.~2024C01020); the Emergency Management Research and Development Project of Zhejiang Province (No.~2024YJ018).

\bibliographystyle{IEEEbib}

\bibliography{refs}

\end{document}

%% file: tables/main_result.tex
\begin{table*}[ht]
\centering
\small
\renewcommand{\arraystretch}{1.2}
\setlength{\tabcolsep}{4pt}
\begin{tabular}{l l cc c cc c cc c cc c | c}
\hline
\multirow{2}{*}{\textbf{Model}} & \multirow{2}{*}{\textbf{Method}} &
\multicolumn{3}{c}{\textbf{AIME24}} &
\multicolumn{3}{c}{\textbf{AIME25}} &
\multicolumn{3}{c}{\textbf{Math500}} &
\multicolumn{3}{c|}{\textbf{AMC23}} &
\multirow{2}{*}{\textbf{Avg. EPR}} \\
& & Acc & Tok & EPR & Acc  & Tok & EPR & Acc  & Tok & EPR & Acc  & Tok & EPR & \\
\hline
\multirow{5}{*}{\textbf{DS-1.5B}}
& Vanilla & 29.2 & 15591 & -- & 25.4 & 15099 & -- & 83.1 & 4984 & -- & 73.0 & 8834 & -- & -- \\
& NOWAIT & 22.1 & 8196 & -- & 18.8 & 7772 & -- & 79.0 & 3086 & -- & 65.3 & 4594 & -- & -- \\
& TIP & 27.9 & 11458 & 3.1 & 22.9 & 9160 & 2.1 & 80.8 & 3419 & 1.5 & 72.0 & 5998 & 5.1 & 2.9 \\
& DEER & 27.5 & 8695(50) & 3.9 & 21.6 & 8807(60) & 1.5 & 70.1 & \textbf{2575(22)} & 0.4 & 68.2 & \textbf{5152(30)} & 1.4 & 1.7 \\
& Entrocut & \textbf{28.8} & \textbf{8295(111)} & \textbf{17.5} & \textbf{23.8} & \textbf{7912(124)} & \textbf{4.0} & \textbf{81.6} & 3341(63) & \textbf{2.4} & \textbf{72.8} & 6000(91) & \textbf{25.7} & \textbf{12.4} \\
\hline
\multirow{5}{*}{\textbf{DS-7B}}
& Vanilla & 55.8 & 12775 & -- & 42.3 & 14319 & -- & 92.1 & 4097 & -- & 90.3 & 6434 & -- & -- \\
& NOWAIT & 46.3 & 7921 & -- & 30.0 & 7449 & -- & 88.7 & 2768 & -- & 82.2 & 3811 & -- & -- \\
& TIP & 51.3 & 9919 & 1.2 & 32.9 & 10138 & 0.8 & 90.5 & 3055 & 1.7 & 87.3 & 4714 & 1.8 & 1.3 \\
& DEER & 49.2 & \textbf{9138(44)} & 1.1 & 37.7 & \textbf{9586(51)} & 1.8 & 89.0 & \textbf{2342(17)} & 1.5 & 86.9 & \textbf{4322(24)} & 1.9 & 1.5 \\
& Entrocut & \textbf{51.7} & 9663(103) & \textbf{1.5} & \textbf{38.1} & 9555(115) & \textbf{2.0} & \textbf{91.1} & 3046(50) & \textbf{2.7} & \textbf{89.2} & 5216(74) & \textbf{3.4} & \textbf{2.4} \\
\hline
\end{tabular}
\caption{Accuracy (Acc), token usage (Tok), Efficiency-Performance Ratio (EPR), and average EPR (Avg EPR) across four datasets for different methods.}
\label{tab:method_comparison_with_epr_compact}
\end{table*}

%% file: tables/ablation_study.tex
% \begin{table}[h]
% \centering
% \begin{tabular}{c c c c c c c c }
% \hline
% \multirow{2}{*}{\textbf{Model}} & \multirow{2}{*}{\textbf{Method}} &
% \multicolumn{3}{c}{\textbf{AIME2024}} &
% \multicolumn{3}{c}{\textbf{AIME2025}} \\
% & & Acc & Tok & EPR & Acc & Tok & EPR \\
% \hline
% \multirow{3}{*}{\textbf{DS-1.5B}} & Entrocut & 28.8 & 8295.8 & 17.51 & 23.8 & 7912.8 & 4.05 \\
% & w/ hard budget & 26.4 & 7535.5 & 2.76 & 22.0 & 7590.8 & 1.99 \\
% & w/o entropy probe & 26.6 & 8263.8 & 2.71 & 21.9 & 8547.1 & 1.69 \\
% \hline
% \multirow{3}{*}{\textbf{DS-7B}} & Entrocut & 51.7 & 9663.3 & 1.47 & 38.1 & 9555.3 & 2.04 \\
% & w/ hard budget & 25.0 & 6437.0 & 0.41 & 17.2 & 6815.8 & 0.54 \\
% & w/o entropy probe & 47.4 & 9096.6 & 0.86 & 37.0 & 9536.6 & 1.62 \\
% \hline
% \end{tabular}
% \caption{EntroCut ablation results on AIME2024/2025 datasets.}
% \label{tab:entrocut_albation}
% \end{table}

% \begin{table}[h]
\begin{table}[!]
\centering
\small
\begin{tabular}{c c c c c}
\hline
\multirow{2}{*}{\textbf{Model}} & \multirow{2}{*}{\textbf{Method}} &
\multicolumn{3}{c }{\textbf{AIME25}} \\
& & Acc & Tok & EPR \\
\hline
\multirow{3}{*}{\textbf{DS-1.5B}} & EntroCut & 23.8 & 7912.8 & 4.05 \\
& w/ hard budget & 22.0 & 7590.8 & 1.99 \\
& w/o entropy probe & 21.9 & 8547.1 & 1.69 \\
\hline
\multirow{3}{*}{\textbf{DS-7B}} & EntroCut & 38.1 & 9555.3 & 2.04 \\
& w/ hard budget & 17.2 & 6815.8 & 0.54 \\
& w/o entropy probe & 37.0 & 9536.6 & 1.62 \\
\hline
\end{tabular}
\caption{EntroCut ablation results on AIME25 datasets.}
\label{tab:entrocut_albation}
\end{table}

%% file: refs.bib
@article{deepseekai2025deepseekr1incentivizingreasoningcapability,
  title = {DeepSeek-R1 incentivizes reasoning in LLMs through reinforcement learning},
  volume = {645},
  url = {http://dx.doi.org/10.1038/s41586-025-09422-z},
  doi = {10.1038/s41586-025-09422-z},
  number = {8081},
  journal = {Nature},
  publisher = {Springer Science and Business Media LLC},
  author = {DeepSeek-AI},
  year = {2025},
  pages = {633--638},
  language = {en}
}

@misc{feng2025efficientreasoningmodelssurvey,
  doi = {10.48550/ARXIV.2504.10903},
  url = {https://arxiv.org/abs/2504.10903},
  author = {Feng,  Sicheng and Fang,  Gongfan and Ma,  Xinyin and Wang,  Xinchao},
  title = {Efficient Reasoning Models: A Survey},
  publisher = {arXiv},
  year = {2025},
  copyright = {Creative Commons Attribution 4.0 International},
  howpublished = {arXiv:2504.10903},
}

@misc{muennighoff2025s1,
  title={s1: Simple test-time scaling},
  author={Muennighoff, Niklas and Yang, Zitong and Shi, Weijia and Li, Xiang Lisa and Fei-Fei, Li and Hajishirzi, Hannaneh and Zettlemoyer, Luke and Liang, Percy and Cand{\`e}s, Emmanuel and Hashimoto, Tatsunori},
  year={2025},
  howpublished = {arXiv:2501.19393},
}

@inproceedings{
yeo2025demystifying,
title={Demystifying Long Chain-of-Thought Reasoning in {LLM}s},
author={Edward Yeo and Yuxuan Tong and Xinyao Niu and Graham Neubig and Xiang Yue},
booktitle={ICLR 2025 Workshop on Navigating and Addressing Data Problems for Foundation Models},
year={2025},
url={https://openreview.net/forum?id=AgtQlhMQ0V}
}

@inproceedings{
aggarwal2025l,
title={L1: Controlling How Long A Reasoning Model Thinks With Reinforcement Learning},
author={Pranjal Aggarwal and Sean Welleck},
booktitle={Second Conference on Language Modeling},
year={2025},
url={https://openreview.net/forum?id=4jdIxXBNve}
}

@inproceedings{ma-etal-2025-cot,
    title = "{C}o{T}-Valve: Length-Compressible Chain-of-Thought Tuning",
    author = "Ma, Xinyin  and
      Wan, Guangnian  and
      Yu, Runpeng  and
      Fang, Gongfan  and
      Wang, Xinchao",
    booktitle = "Proceedings of the 63rd Annual Meeting of the Association for Computational Linguistics (Volume 1: Long Papers)",
    month = jul,
    year = "2025",
    address = "Vienna, Austria",
    url = "https://aclanthology.org/2025.acl-long.300/",
    doi = "10.18653/v1/2025.acl-long.300",
    pages = "6025--6035",
    ISBN = "979-8-89176-251-0"
}

@inproceedings{
liu2024can,
title={Can Language Models Learn to Skip Steps?},
author={Tengxiao Liu and Qipeng Guo and Xiangkun Hu and Cheng Jiayang and Yue Zhang and Xipeng Qiu and Zheng Zhang},
booktitle={The Thirty-eighth Annual Conference on Neural Information Processing Systems},
year={2024},
url={https://openreview.net/forum?id=w4AnTVxAO9}
}

@inproceedings{nowait,
    title = "Wait, We Don{'}t Need to ``Wait''! Removing Thinking Tokens Improves Reasoning Efficiency",
    author = "Wang, Chenlong  and
      Feng, Yuanning  and
      Chen, Dongping  and
      Chu, Zhaoyang  and
      Krishna, Ranjay  and
      Zhou, Tianyi",
    booktitle = "Findings of the Association for Computational Linguistics: EMNLP 2025",
    month = nov,
    year = "2025",
    address = "Suzhou, China",
    url = "https://aclanthology.org/2025.findings-emnlp.394/",
    doi = "10.18653/v1/2025.findings-emnlp.394",
    pages = "7459--7482",
    ISBN = "979-8-89176-335-7"
}

@misc{tip,
      title={Thoughts Are All Over the Place: On the Underthinking of o1-Like LLMs}, 
      author={Yue Wang and Qiuzhi Liu and Jiahao Xu and Tian Liang and Xingyu Chen and Zhiwei He and Linfeng Song and Dian Yu and Juntao Li and Zhuosheng Zhang and others},
      year={2025},
      eprint={2501.18585},
      archivePrefix={arXiv},
      primaryClass={cs.CL},
      url={https://arxiv.org/abs/2501.18585}, 
      howpublished = {arXiv:2501.18585},
}

@misc{deer,
      title={Dynamic Early Exit in Reasoning Models}, 
      author={Chenxu Yang and Qingyi Si and Yongjie Duan and Zheliang Zhu and Chenyu Zhu and Qiaowei Li and Zheng Lin and Li Cao and Weiping Wang},
      year={2025},
      eprint={2504.15895},
      archivePrefix={arXiv},
      primaryClass={cs.CL},
      url={https://arxiv.org/abs/2504.15895}, 
      howpublished = {arXiv:2504.15895},
}

@inproceedings{
    hendrycks2021measuring,
    title={Measuring Mathematical Problem Solving With the {MATH} Dataset},
    author={Dan Hendrycks and Collin Burns and Saurav Kadavath and Akul Arora and Steven Basart and Eric Tang and Dawn Song and Jacob Steinhardt},
    booktitle={Thirty-fifth Conference on Neural Information Processing Systems Datasets and Benchmarks Track (Round 2)},
    year={2021},
    url={https://openreview.net/forum?id=7Bywt2mQsCe}
}

@misc{aime,
    title={Aime problems and solutions}, 
    author={AoPS},
    howpublished = {Available at \url{https://artofproblemsolving.com/wiki/index.php/AIME_Problems_and_Solutions}}
}

@misc{amc,
    title={Amc 2023, 2024}, 
    author={AI-MO},
    howpublished = {Available at \url{https://huggingface.co/datasets/AI-MO/aimo-validation-amc}}
}

@article{li202512surveyreasoning,
  author={Zhang, Duzhen and Li, Zhong-Zhi and Zhang, Ming-Liang and Zhang, Jiaxin and Liu, Zengyan and Yao, Yuxuan and Xu, Haotian and Zheng, Junhao and Chen, Xiuyi and Zhang, Yingying and Yin, Fei and Dong, Jiahua and Guo, Zhijiang and Song, Le and Liu, Cheng-Lin},
  journal={IEEE Transactions on Pattern Analysis and Machine Intelligence}, 
  title={From System 1 to System 2: A Survey of Reasoning Large Language Models}, 
  year={2025},
  volume={},
  number={},
  pages={1-20},
  doi={10.1109/TPAMI.2025.3637037}}

@misc{sui2025stopoverthinkingsurveyefficient,
      title={Stop Overthinking: A Survey on Efficient Reasoning for Large Language Models}, 
      author={Yang Sui and Yu-Neng Chuang and Guanchu Wang and Jiamu Zhang and Tianyi Zhang and Jiayi Yuan and Hongyi Liu and Andrew Wen and Shaochen Zhong and Na Zou and Hanjie Chen and Xia Hu},
      year={2025},
      eprint={2503.16419},
      archivePrefix={arXiv},
      primaryClass={cs.CL},
      url={https://arxiv.org/abs/2503.16419}, 
      howpublished={arXiv:2503.16419},
}

@article{wei2022chain,
  title={Chain-of-thought prompting elicits reasoning in large language models},
  author={Wei, Jason and Wang, Xuezhi and Schuurmans, Dale and Bosma, Maarten and Xia, Fei and Chi, Ed and Le, Quoc V and Zhou, Denny and others},
  journal={Advances in neural information processing systems},
  volume={35},
  pages={24824--24837},
  year={2022}
}

@inproceedings{han-etal-2025-token,
    title = "Token-Budget-Aware {LLM} Reasoning",
    author = "Han, Tingxu  and
      Wang, Zhenting  and
      Fang, Chunrong  and
      Zhao, Shiyu  and
      Ma, Shiqing  and
      Chen, Zhenyu",
    booktitle = "Findings of the Association for Computational Linguistics: ACL 2025",
    address = "Vienna, Austria",
    url = "https://aclanthology.org/2025.findings-acl.1274/",
    doi = "10.18653/v1/2025.findings-acl.1274",
    pages = "24842--24855",
    ISBN = "979-8-89176-256-5",
    month = jul,
    year = "2025",
}

@inproceedings{
luo2025opruner,
title={O1-Pruner: Length-Harmonizing Fine-Tuning for O1-Like Reasoning Pruning},
author={Haotian Luo and Li Shen and Haiying He and Yibo Wang and Shiwei Liu and Wei Li and Naiqiang Tan and Xiaochun Cao and Dacheng Tao},
booktitle={2nd AI for Math Workshop @ ICML 2025},
year={2025},
url={https://openreview.net/forum?id=ioYybCRcyW}
}
